\documentclass[10pt,twocolumn,letterpaper]{article}

\usepackage{iccv}
\usepackage{times}
\usepackage{epsfig}
\usepackage{graphicx}
\usepackage{amsmath}
\usepackage{amssymb}

\usepackage[pagebackref=true,breaklinks=true,letterpaper=true,colorlinks,bookmarks=false]{hyperref}

\iccvfinalcopy 

\def\iccvPaperID{****} 

\ificcvfinal\fi

\begin{document}

\title{Uncertainty-Aware Anticipation of Activities}

\author{Yazan Abu Farha and Juergen Gall\\
University of Bonn, Germany\\
{\tt\small \{abufarha,gall\}@iai.uni-bonn.de}
}


\maketitle
\ificcvfinal\thispagestyle{empty}\fi

\begin{abstract}
Anticipating future activities in video is a task with many practical applications. While 
earlier approaches are limited to just a few seconds in the future, the prediction time 
horizon has just recently been extended to several minutes in the future. However, 
as increasing the predicted time horizon, the future becomes more uncertain and models 
that generate a single prediction fail at capturing the different possible future activities. 
In this paper, we address the uncertainty modelling for predicting long-term future activities. 
Both an action model and a length model are trained to model the probability distribution of the 
future activities. At test time, we sample from the predicted distributions multiple samples that 
correspond to the different possible sequences of future activities. Our model is evaluated on 
two challenging datasets and shows a good performance in capturing the multi-modal future 
activities without compromising the accuracy when predicting a single sequence of future activities.
\end{abstract}

\section{Introduction}
\label{sec:intro}
Anticipating future activities in video has become an active research topic in computer vision. 
While earlier approaches focused on early activity 
detection~\cite{ryoo2011human,hoai2014max,ma2016learning,sadegh2017encouraging}, 
recent models predict activities a few seconds in the 
future~\cite{lan2014hierarchical,jain2016structural,vondrick2016anticipating,gao2017red}. 
However, predicting the future activity label shortly before it starts is not sufficient for 
many applications. Robots, for instance, that interact with humans to accomplish industrial 
tasks or help in housework need to anticipate activities for a long time horizon. Such long-term 
prediction would enable these robots to plan ahead to complete their tasks efficiently. Moreover, 
anticipating the activities of other interacting humans improves human-robot interaction. 

\begin{figure*}[t]
\begin{center}
\vspace{.25cm}
   \includegraphics[width=.95\linewidth]{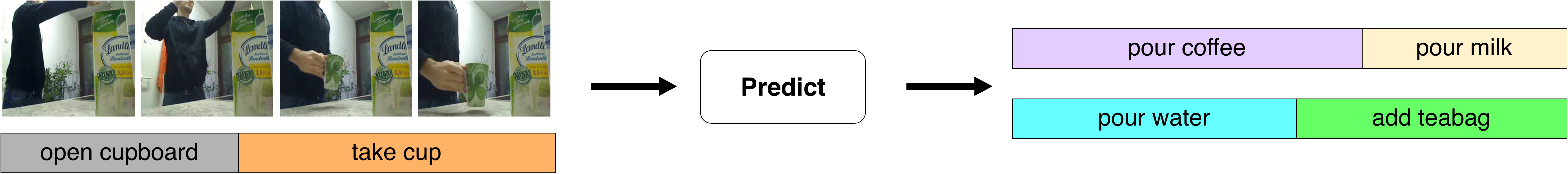}
\end{center}
   \caption{The anticipation task. Given an observed part of a video, we want to predict 
   the future activities that might follow in the future with their durations. The model outputs 
   a collection of samples to represent the uncertainty in the future actions.}
\label{fig:intro}
\end{figure*}

Recently, \cite{abufarha2018when} extended the prediction horizon to a few minutes in the future and 
predict both future activities and their durations as well. While their approach generates good 
predictions, it does not take the uncertainty of the future into consideration. For example, 
given a video snippet that shows a person taking a cup from the cupboard, we cannot be sure 
whether the future activity would be pour water or pour coffee. Approaches that 
predict a single output and do not model the uncertainty in the future actions would fail in such cases. 
On the contrary, approaches that are capable of predicting all the possible outputs are preferable. 
Fig.~\ref{fig:intro} illustrates a case where the future activities have multiple modes and the model 
has to predict all these modes. 

In this paper, we introduce a framework that models the probability distribution of the 
future activities and use this distribution to generate several possible sequences of future 
activities at test time. To this end, we train an action model that predicts a probability distribution 
of the future action label, and a length model which predicts a probability distribution of the future action 
length. At test time, we sample from these models a future action segment represented by an action 
label and its length. To predict more in the future, we feed the predicted action segment to the model 
and predict the next one recursively. We evaluate our framework on two datasets with videos of varying 
length and many action segments: the Breakfast dataset~\cite{kuehne2014language} and 
50Salads~\cite{stein2013combining}. Our framework outperforms a baseline that predicts the future 
activities using n-grams as an action model and a Gaussian to predict the action length. Furthermore, 
we are able to achieve results that are comparable with the state-of-the-art if we use the framework to 
predict a single sequence of future activities.

\section{Related Work}
\label{sec:related_work}
Future prediction has been studied by many researchers. However, the predicted time horizon 
is very limited in earlier approaches. Instead of predicting the future, Hoai and 
De la Torre~\cite{hoai2014max} proposed a max-margin framework for early activity detection. 
Other approaches adapt special loss functions to detect a partially observed 
activity~\cite{ma2016learning,sadegh2017encouraging}. To predict future actions, 
Lan~\etal~\cite{lan2014hierarchical} proposed hierarchical representations of short clips 
or still images. In~\cite{koppula2016anticipating} a spatio-temporal graph is used to predict 
object affordances, trajectories, and sub-activities. Vondrick~\etal~\cite{vondrick2016anticipating} 
trained a deep convolutional neural network to predict features in the future from a single frame. 
An SVM is then used to predict the future action label from the predicted features. In~\cite{gao2017red}, 
a sequence of visual representations of past frames is used to predict a sequence of future representations. 
A reinforcement learning module is used to provide supervision at sequence level. Zeng~\etal~\cite{zeng2017visual} 
used inverse reinforcement learning to anticipate visual representations from unlabeled video.
Shi~\etal~\cite{shi2018action} proposed a recurrent neural network with radial basis function kernels 
to predict features in the future and then predict the action label with a multi-layer perceptron. 
Instead of features, \cite{rodriguez2018action} predict future dynamic images and train a classifier 
to predict the action label on top of the predicted images. Furnari~\etal~\cite{furnari2018leveraging} 
predict future actions in egocentric videos and evaluate the top-k accuracy to consider the multi-modal 
future. Miech~\etal~\cite{miech2019leveraging} combine a predictive model that directly predicts the future 
action label with a  transitional model that models the transition probabilities between actions. 
However, all of the previous approaches predict an action label without any time information. 
Heidarivincheh~\etal~\cite{heidarivincheh2018action} introduced a model to predict the time of 
completion for an observed activity. In~\cite{mahmud2017joint}, both the future activity 
and its starting time are predicted.

Despite the success of the previous approaches in predicting the future actions, they are however 
limited to a few seconds in the future. For many real world applications, a long-term prediction 
beyond just a few seconds is crucial. Recently, Abu Farha~\etal~\cite{abufarha2018when} introduced 
a two-step approach that is capable of anticipating future activities several minutes in the future. 
Given an observed part of a video, they infer the activities in the observed part first and 
then anticipate the future activities and their durations. For the anticipation step, both 
an RNN and a CNN are trained to generate the future action segments. Ke~\etal~\cite{ke2019time} build on 
this two-step framework and use temporal convolutions with attention to anticipate future activities. The model output is conditioned on a time parameter to determine the predicted time horizon.
In contrast to these approaches, we predict multiple outputs to consider the 
uncertainty in the future activities. In a very recent work,~\cite{mehrasa2019variational} adapt 
a variational auto-encoder framework to predict a distribution over the future action and its starting 
time. However, they do not model the dependency between the future action and its starting time directly 
and rely on a shared latent space to capture this dependency. On the contrary, our framework directly 
models the dependency between the action label and its length.

\section{Anticipating Activities}
\label{sec:anticipating_activities}
Given an observed part of a video with $n$ action segments \mbox{$c_{1:n} = (c_1, \dots, c_n)$} with 
length $l_{1:n} = (l_1, \dots, l_n)$, we want to predict all the action segments and their lengths that 
will occur in the future unseen part of that video. \Ie~we want to predict the segments $c_{n+1:N}$ and 
the corresponding segments length $l_{n+1:N}$, where $N$ is the total number of action segments in the 
video. Furthermore, since the last observed action segment $c_{n}$ might be partially observed and 
will continue in the future, we want to update our estimate of $l_{n}$ as well. Since for 
the same observed action segments $c_{1:n}$ there are more than one possible future action segments, 
we want to predict more than one output to capture the different modes in the future as shown in 
Fig.~\ref{fig:intro}. To this end, we propose a framework to model the uncertainty in the future 
activities, and then use this framework to generate samples of the future action segments. We start with 
the model description in Section~\ref{sec:model}, and then describe the prediction procedure in 
Section~\ref{sec:prediction}.

\subsection{Model}
\label{sec:model}

\begin{figure*}[t]
\vspace{2mm}
\begin{center}
\begin{tabular}{p{2cm} c p{2cm} c}
   \multicolumn{4}{c}{\includegraphics[trim={0cm 0cm 0cm 0cm},clip,width=\linewidth]{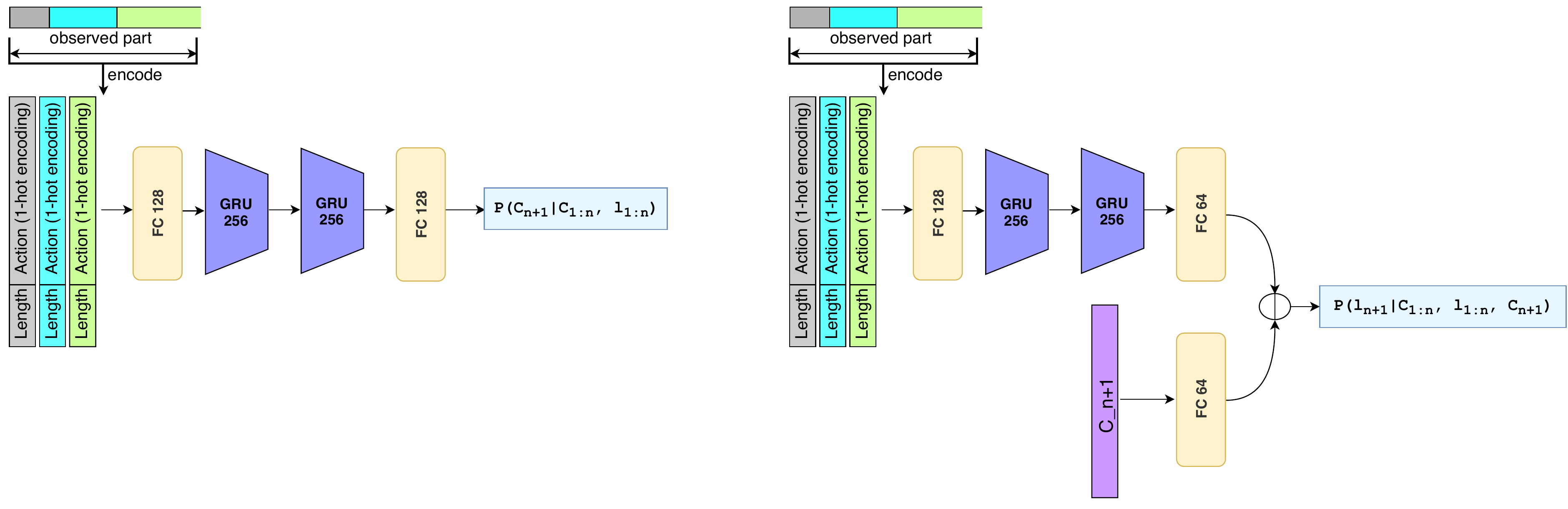}}  
   \\ 
   &  & (a)&  (b) 
\end{tabular}
\end{center}
\vspace{1mm}
   \caption{Our anticipation framework consists of: (a) The action model which predicts the probability 
   distribution of the future action label. (b) The length model which predicts the probability distribution 
   of the future action length.}
\label{fig:models}
\end{figure*}

We follow the two-step approach proposed in~\cite{abufarha2018when} and infer 
the actions in the observed frames and then predict the future actions. For 
inferring the actions in the observed part, we use the same RNN-HMM model~\cite{richard2017weakly} 
that is used by~\cite{abufarha2018when}. Our goal now is to model the probability 
of the future actions and their lengths. This can be done using an autoregressive model 
that predicts the future action and its length, and then feeding the predicted output to the 
model again to predict the next one. Using such an autoregressive approach allows us to use the 
same model to predict actions for arbitrarily long time horizons. The probability distribution of the 
future action segment and its length can be factorized as follows

\begin{equation}
\begin{split}
p(c_{n+1},\ l_{n+1} | c_{1:n},\ l_{1:n}) = 
p(c_{n+1} | c_{1:n},\ l_{1:n}) \ \cdot \\
p(l_{n+1} | c_{1:n},\ l_{1:n},\ c_{n+1}),
\end{split}
\end{equation}
where the first factor $p(c_{n+1} | c_{1:n},\ l_{1:n})$ is an action model that describes 
the probability distribution of the future action label given the preceding action segments. 
Whereas the second factor $p(l_{n+1} | c_{1:n},\ l_{1:n},\ c_{n+1})$ is a length model that describes 
the probability distribution of the future action length given the preceding action segments and the 
future action label. In the following, we discuss the details of these models. 

\subsubsection{Action Model}
\label{sec:action_model}
The action model predicts a probability distribution of the future action given 
the sequence of observed action segments and their lengths. For this model we use 
an RNN-based model similar to the one proposed in~\cite{abufarha2018when} as shown in 
Fig.~\ref{fig:models} (a). Given the observed part of the video, each action 
segment is represented by a vector of a 1-hot encoding of the action label and 
the corresponding segment length. This sequence is passed through a fully connected 
layer, two layers of gated recurrent units (GRUs), and another fully connected layer. 
We use ReLU activations for the fully connected layers. For the output layer, we use 
another fully connected layer that predicts action scores for the future action. 
To get the probability distribution of the future action, we apply a softmax over the 
predicted scores
\begin{equation}
p(c_{n+1}=\hat{c} | c_{1:n},\ l_{1:n}) = \frac{e^{x_{\hat{c}}}}{\sum_{\tilde{c}}e^{x_{\tilde{c}}}},
\end{equation}
where $x_{\hat{c}}$ is the predicted action score for the class $\hat{c}$. To train this model, 
we use a cross entropy loss
\begin{equation}
\mathcal{L}_{action} = \frac{1}{M}\sum_{m} -log(y_{m,c}), 
\end{equation}
where $y_{m,c}$ is the predicted probability for the ground-truth future action label $c$ 
in the $m^{th}$ training example.

\subsubsection{Length Model}
\label{sec:length_model}
We model the probability distribution of the future action length with a Gaussian distribution, \ie
\begin{equation}
p(l_{n+1} | c_{1:n},\ l_{1:n},\ c_{n+1}) = \mathcal{N}(\mu,\ \sigma^{2}).
\end{equation}
To predict the mean length $\mu$ and the standard deviation $\sigma$ of this distribution, we use 
a network with two branches as shown in Fig.~\ref{fig:models} (b). The first branch is the same 
as the RNN-based model used in the action model. It takes the sequence of the observed action segments 
as input and encodes them into a single vector representation. Whereas the second branch is a single 
fully connected layer that takes a 1-hot encoding of the future action $c_{n+1}$ and encodes it in 
another single vector representation. These two encoded vectors are concatenated and passed through 
two fully connected output layers to predict the mean length $\mu$ and the standard deviation $\sigma$. 
To ensure that the standard deviation is non-negative we use exponential activations for the output layer. 
To train the length model, we minimize the negative log likelihood of the target length 

\begin{equation}
\begin{split}
\mathcal{L}_{length} &= \frac{1}{M}\sum_{m} - log \ p(l_{n+1} =\ell_m|c_{1:n},\ l_{1:n},\ c_{n+1}), \\
&= \frac{1}{M}\sum_{m} 0.5 \ log(2 \pi) \ + log(\sigma_{m}) \ + \\
&\quad \ \ \frac{(\ell_m \ - \ \mu_{m})^{2}}{2 \sigma_{m}^{2}},
\end{split}
\label{eqn:len_loss}
\end{equation}
where $\ell_m$ is the ground-truth length of the future action, $\mu_m$ and $\sigma_m$ are the 
predicted mean length and standard deviation for the $m^{th}$ training example. As the first term in 
(\ref{eqn:len_loss}) is constant, our final loss function for the length model can be reduced to 
\begin{equation}
\mathcal{L}_{length} =  \frac{1}{M}\sum_{m} log(\sigma_{m}) \ + \frac{(\ell_m \ - \ \mu_{m})^{2}}{2 \sigma_{m}^{2}}.
\end{equation}

The training examples for both the action and length models are generated based on 
the ground-truth segmentation of the training videos. Given a video with $n$ action 
segments, we generate $n-1$ training examples. For a segment $i > 1$, the sequence of 
all the preceding segments is considered as input of the training example, where each segment 
is represented by a vector of a 1-hot encoding of the action label and the corresponding segment 
length. The action label of segment $i$ defines the target for the action model and its length 
serves as a target for the length model.

\subsection{Prediction}
\label{sec:prediction}
Given an observed part of a video with $n$ action segments, we want to generate plausible sequences of future 
activities. Note that the observed part might end in a middle of an action segment and the last segment 
in the observations might be not fully observed. In the following we describe two strategies for predicting the 
sequence of future activities. The first strategy generates multiple sequences of activities by sampling from the 
predicted distributions of our approach. Whereas the second strategy is used to generate a single prediction 
that corresponds to the mode of the predicted distributions.

\subsubsection{Prediction by Generating Samples}
At test time we alternate between two steps: sampling a future action label 
from the action model 
\begin{equation}
\hat{c}_{n+1} \sim p(c_{n+1} | c_{1:n},\ l_{1:n}),
\end{equation} 
and then we sample a length for the sampled future action using the length model
\begin{equation}
\hat{l}_{n+1} \sim p(l_{n+1} | c_{1:n},\ l_{1:n},\ \hat{c}_{n+1}).
\end{equation} 
We feed the predicted action segment recursively to the model until we predict the desired 
time horizon. As the last observed action segment might continue in the future, we start 
the prediction with updating the length of the last observed action segment. For this step, 
we sample a length for the last observed action segment based on the preceding segments and 
only update the length of that segment if the generated sample is greater than 
the observed length as follows
\begin{equation}
\hat{l}_{n} \sim p(l_{n} | c_{1:n-1},\ l_{1:n-1},\ c_{n}), 
\end{equation} 
\begin{equation}
l_{n}^{\star} = 
\begin{cases}
\hat{l}_{n}        &: \hat{l}_{n}  > \ell_{n}\\
\ell_{n}                &: otherwise\\
\end{cases}, 
\end{equation}
where $\ell_{n}$ is the observed length of the last observed action segment, 
and $l_{n}^{\star}$ is the predicted full length of that segment.

\subsubsection{Prediction of the Mode}
For predicting the mode, we also alternate between predicting the future action label and then 
predicting the length of that label. However, instead of sampling from the predicted action distribution, 
we choose the action label with the highest probability. For predicting the length, we use 
the predicted mean length from the length model instead of sampling from the predicted distribution.

\subsection{Implementation Details}
\label{sec:implementation_details}
We implemented both models in PyTorch~\cite{paszke2017automatic} and trained them using 
Adam optimizer~\cite{kingma2015adam} with a learning rate of $0.001$. The batch size 
is set to $32$. We trained the action model for $60$ epochs and the length model is 
trained for $30$ epochs. Dropout is used after each layer with probability $0.5$. 
We also apply standardization to the length of the action segments as follows
\begin{equation}
l = \frac{l - \overline{l}}{\sigma_{l}},
\end{equation}
where $\overline{l}$ and $\sigma_{l}$ are the mean length and standard deviation, respectively, 
computed from the lengths of all action segments in the training videos.

\section{Experiments}
\label{sec:experiments}

\paragraph{\textbf{Datasets.}} We evaluate the proposed model on two challenging 
datasets: the Breakfast dataset~\cite{kuehne2014language} and 50Salads~\cite{stein2013combining}.

The \textbf{Breakfast} dataset contains $1,712$ videos with roughly $3.6$ million frames. 
Each video belongs to one out of ten breakfast related activities, such as make tea or pancakes. 
The video frames are annotated with fine-grained action labels like pour water or take cup. Overall, 
there are $48$ different actions where each video contains $6$ action instances on 
average. The videos were recorded by $52$ actors in 18 different kitchens with varying view points. 
For evaluation, we use the standard 4 splits as proposed in~\cite{kuehne2014language} 
and report the average.

The \textbf{50Salads} dataset contains 50 videos with roughly $600,000$ frames. 
On average, each video contains $20$ action instances and is $6.4$ minutes long. 
All the videos correspond to salad preparation activities and were performed by $25$ actors. 
The video frames are annotated with $17$ different fine-grained action labels like
cut tomato or peel cucumber. For evaluation, we use five-fold cross-validation and 
report the average as in~\cite{stein2013combining}. 

\paragraph{\textbf{Evaluation Metric.}} We evaluate our framework using two evaluation protocols. 
The first protocol is used in~\cite{abufarha2018when} where we observe $20\%$ or $30\%$ of the 
video and predict the following $10\%,\ 20\%,\ 30\%$ and $50\%$ of that video. As a metric, we report 
the mean over classes (MoC) by evaluating the frame-wise accuracy of each action class and then averaging 
over the total number of ground-truth action classes. To evaluate multiple samples of future activities, 
the average frame-wise accuracy of each action class is used to compute the MoC. The average accuracy over 
samples has been used in other future prediction tasks like predicting future semantic 
segmentation~\cite{bhattacharyya2018bayesian} or predicting the next future action label~\cite{mehrasa2019variational}.
The second protocol is used in~\cite{mehrasa2019variational} where we predict only the next future action 
segment and report the accuracy of the predicted label.

\paragraph{\textbf{Baseline.}} As a baseline we replace our action model with tri-grams, 
where the probability of the action is determined based on the preceding two action 
segments. For the length model, we assume the length of an action follows a Gaussian 
distribution with the mean and variance estimated from the training action segments of 
the corresponding action.

\begin{table}[tb]
\centering
\resizebox{\columnwidth}{!}{%
\begin{tabular}{|c|c|c|c|c|c|c|c|c|}
\hline
Observation \% & 
\multicolumn{4}{c|}{\textbf{20\%}} &
\multicolumn{4}{c|}{\textbf{30\%}} \\ \hline
Prediction \%  & 
\textbf{10\%} & \textbf{20\%} & \textbf{30\%} & \textbf{50\%} & 
\textbf{10\%} & \textbf{20\%} & \textbf{30\%} & \textbf{50\%}  \\ \hline

\multicolumn{9}{|l|}{\textbf{\textit{Breakfast}}} \\ \hline
bi-grams &
0.4511      &       0.3503       &       0.3094      &       0.2569      &
0.4578      &       0.3629       &       0.3148      &       0.2612    
\\ 
tri-grams &
0.4595      &       0.3759       &       0.3413      &       0.2952      &
0.4809      &       0.4030       &       0.3586      &       0.3060    
\\ 
four-grams &
0.4728      &       0.3855       &       0.3474      &       0.2970      &
0.4988      &       0.4143       &       0.3645      &       0.3086    
\\ 

Ours      & 
\textbf{0.5039}      &     \textbf{0.4171}       &     \textbf{0.3779}      &       \textbf{0.3278}      &
\textbf{0.5125}      &     \textbf{0.4294}       &     \textbf{0.3833}      &       \textbf{0.3307}         
\\ \hline

\multicolumn{9}{|l|}{\textbf{\textit{50Salads}}} \\ \hline
bi-grams &
0.3039      &       0.2230       &       0.1853      &       0.1075      &
0.3153      &       0.1859       &       0.1295      &       0.0884    
\\ 
tri-grams &
0.3188      &       0.2313       &       0.1919      &       0.1158      &
0.3207      &       0.1931       &       0.1390      &       0.0940  
\\ 
four-grams &
0.3042      &       0.2253       &       0.1831      &       0.1125      &
0.3079      &       0.1889       &       0.1309      &       0.0958    
\\ 

Ours      & 
\textbf{0.3495}      &     \textbf{0.2805}       &     \textbf{0.2408}      &       \textbf{0.1541}      &
\textbf{0.3315}      &     \textbf{0.2465}       &     \textbf{0.1884}      &       \textbf{0.1434}  
\\ \hline

\end{tabular}%
}
\vspace{1mm}
\caption{Results for anticipation with ground-truth observations. Numbers represent mean over classes (MoC) 
metric averaged over $25$ samples.}
\label{tab:avg_gt}
\end{table}

\begin{table}[tb]
\centering
\resizebox{\columnwidth}{!}{%
\begin{tabular}{|c|c|c|c|c|c|c|c|c|}
\hline
Observation \% & 
\multicolumn{4}{c|}{\textbf{20\%}} &
\multicolumn{4}{c|}{\textbf{30\%}} \\ \hline
Prediction \%  & 
\textbf{10\%} & \textbf{20\%} & \textbf{30\%} & \textbf{50\%} & 
\textbf{10\%} & \textbf{20\%} & \textbf{30\%} & \textbf{50\%}  \\ \hline

\multicolumn{9}{|l|}{\textbf{\textit{Breakfast}}} \\ \hline
Baseline &
0.4954      &       0.4038       &       0.3683      &       0.3271      &
0.5225      &       0.4295       &       0.3892      &       0.3380  
\\ 

Ours      & 
\textbf{0.5300}      &       \textbf{0.4410}       &       \textbf{0.3972}      &       \textbf{0.3490}      &
\textbf{0.5399}      &       \textbf{0.4453}       &       \textbf{0.4021}      &       \textbf{0.3558}     
\\ \hline

\multicolumn{9}{|l|}{\textbf{\textit{50Salads}}} \\ \hline
Baseline &
0.3165      &       0.2502       &       0.2075      &       0.1241      &
0.3698      &       0.2251       &       0.1604      &       0.1158  
\\ 

Ours      & 
\textbf{0.3810}      &       \textbf{0.3010}       &       \textbf{0.2633}      &       \textbf{0.1651}      &
\textbf{0.4000}     &       \textbf{0.2927}       &       \textbf{0.2317}      &       \textbf{0.1548}        
\\ \hline

\end{tabular}%
}
\vspace{1mm}
\caption{Results for anticipation with ground-truth observations. Numbers represent mean over classes (MoC) 
metric of the predicted distribution mode.}
\label{tab:argmax_gt}

\end{table}

\subsection{Anticipation with Ground-Truth Observations}

We start the evaluation by using the ground-truth annotations of the observed part. 
This setup allows us to isolate the effect of the action segmentation model which 
is used to infer the labels of the observed part (\ie~the RNN-HMM model). Table~\ref{tab:avg_gt} 
shows the results of our model compared to n-grams baselines. For each example in the test set, we 
generate $25$ samples and use the average accuracy of each class to compute the mean over 
classes metric (MoC). As shown in Table~\ref{tab:avg_gt}, our approach outperforms the baselines on both 
datasets and in all the test cases. This indicates that our model learns a better distribution 
of the future action segments represented by the generated samples. We also show the effect of 
using bi-grams or four-grams instead of the used tri-grams for the baseline. As shown in Table~\ref{tab:avg_gt}, 
the effect of using different n-grams model is small. While using more history gives a slight 
improvement on the Breakfast dataset, but the tri-grams model performs better than four-grams on the 
50Salads dataset, which contains much longer sequences. For the rest of the experiments, we stick with the 
tri-grams model for the proposed baseline.

We also report the accuracy of the mode of the 
predicted distribution. Instead of randomly drawing a sample from the predicted distribution, we predict 
the action label with the highest probability at each step. For the action length, we use the predicted 
mean length. Table~\ref{tab:argmax_gt} shows the results on both 50Salads and the Breakfast dataset. 
Our approach outperforms the baseline in this setup as well.

\begin{figure*}[tb]
\begin{center}
\vspace{.25cm}
   \includegraphics[trim={1.45cm 0.2cm 1cm .25cm},clip,width=.8\linewidth]{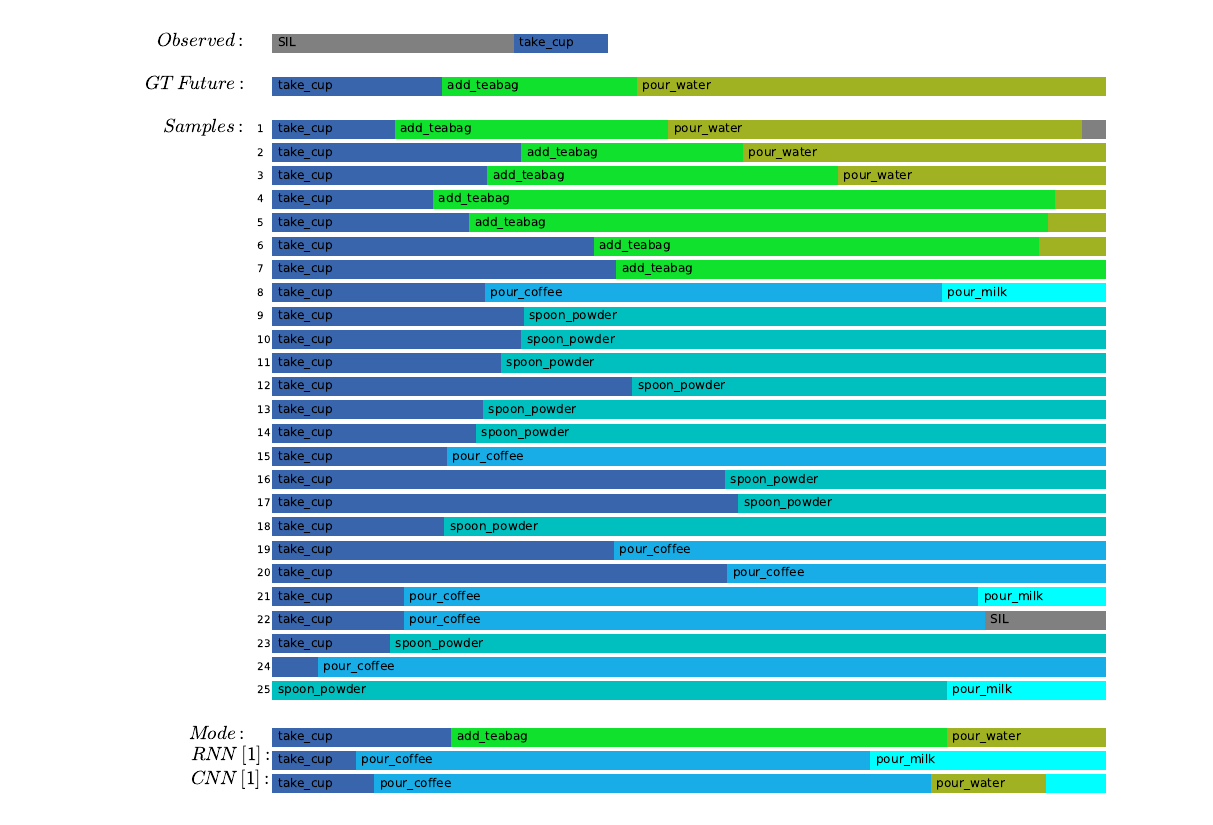}
\end{center}
\vspace{-3mm}
\caption{Qualitative result from the Breakfast dataset. This sequence corresponds to the activity 
of {\tt{making\_tea}}. We observe $20\%$ of the video and predict the following $50\%$. Both 
the generated samples and the mode of the predicted distribution are shown. The samples are ranked 
based on the frame-wise accuracy of the predicted activities. We also show the results of the RNN and CNN models from \cite{abufarha2018when}.}
\label{fig:q_res}
\end{figure*}

Fig.~\ref{fig:q_res} shows a qualitative result from the Breakfast dataset. Both the generated samples 
and the mode of the distribution are shown. We also show the results of the RNN and CNN models 
from \cite{abufarha2018when}. As illustrated in the figure, there are many possible action segments that 
might happen after the observed part ({\tt{SIL, take\_cup}}), and our model is able to generate samples 
that correspond to these different possibilities. In contrast, \cite{abufarha2018when} generates only one 
possible future sequence of activities, which in this case does not correspond to the ground-truth.

\subsection{Anticipation without Ground-Truth Observations}
\label{sec:anticipation_without_gt_obs}

\begin{table}[tb]
\centering
\resizebox{\columnwidth}{!}{%
\begin{tabular}{|c|c|c|c|c|c|c|c|c|}
\hline
Observation \% & 
\multicolumn{4}{c|}{\textbf{20\%}} &
\multicolumn{4}{c|}{\textbf{30\%}} \\ \hline
Prediction \%  & 
\textbf{10\%} & \textbf{20\%} & \textbf{30\%} & \textbf{50\%} & 
\textbf{10\%} & \textbf{20\%} & \textbf{30\%} & \textbf{50\%}  \\ \hline

\multicolumn{9}{|l|}{\textbf{\textit{Breakfast}}} \\ \hline
Baseline &
0.1539      &       0.1365       &       0.1293      &       0.1190      &
\textbf{0.1931}      &       0.1656       &       0.1576      &       0.1390    
\\ 

Ours      & 
\textbf{0.1569}      &       \textbf{0.1400}       &       \textbf{0.1330}      &       \textbf{0.1295}      &
0.1914      &       \textbf{0.1718}       &       \textbf{0.1738}      &       \textbf{0.1498}         
\\ \hline

\multicolumn{9}{|l|}{\textbf{\textit{50Salads}}} \\ \hline
Baseline &
0.2141      &       0.1636        &       0.1329      &       0.0939      &
0.2459      &       0.1560        &       0.1173      &       0.0857   
\\ 

Ours      & 
\textbf{0.2356}      &  \textbf{0.1948}       &   \textbf{0.1801}      &       \textbf{0.1278}      &
\textbf{0.2804}      &  \textbf{0.1795}       &   \textbf{0.1477}      &       \textbf{0.1206}  
\\ \hline

\end{tabular}%
}
\vspace{1mm}
\caption{Results for anticipation without ground-truth observations. Numbers represent mean over classes (MoC) 
metric averaged over $25$ samples.}
\label{tab:avg_noisy}
\end{table}

\begin{table}[tb]
\centering
\resizebox{\columnwidth}{!}{%
\begin{tabular}{|c|c|c|c|c|c|c|c|c|}
\hline
Observation \% & 
\multicolumn{4}{c|}{\textbf{20\%}} &
\multicolumn{4}{c|}{\textbf{30\%}} \\ \hline
Prediction \%  & 
\textbf{10\%} & \textbf{20\%} & \textbf{30\%} & \textbf{50\%} & 
\textbf{10\%} & \textbf{20\%} & \textbf{30\%} & \textbf{50\%}  \\ \hline

\multicolumn{9}{|l|}{\textbf{\textit{Breakfast}}} \\ \hline
Baseline &
0.1655       &       0.1474       &       0.1385      &       0.1319       &
\textbf{0.2080}       &       0.1767       &       0.1700      &       0.1578   	
\\ 

Ours      & 
\textbf{0.1671}       &       \textbf{0.1540}       &       \textbf{0.1447}      &       \textbf{0.1420}      &
0.2073       &       \textbf{0.1827}       &       \textbf{0.1842}      &       \textbf{0.1686}        
\\ \hline

\multicolumn{9}{|l|}{\textbf{\textit{50Salads}}} \\ \hline
Baseline &
0.2174       &       0.1743       &       0.1496      &       0.1034       &
0.2806       &       0.1870       &       0.1460      &       0.0975     
\\ 

Ours      & 
\textbf{0.2486}       &       \textbf{0.2237}       &       \textbf{0.1988}      &       \textbf{0.1282}      &
\textbf{0.2910}       &       \textbf{0.2050}       &       \textbf{0.1528}      &       \textbf{0.1231}        
\\ \hline

\end{tabular}%
}
\vspace{1mm}
\caption{Results for anticipation without ground-truth observations. Numbers represent mean over classes (MoC) 
metric of the predicted distribution mode.}
\label{tab:argmax_noisy}

\end{table}

In this section, we evaluate our approach without relying on the ground-truth annotations of the 
observed part. \Ie~we infer the labels of the observed part of the video with the RNN-HMM 
model~\cite{richard2017weakly}, and then use our approach to predict the future activities. 
Table~\ref{tab:avg_noisy} reports the results of the generated samples, represented by the mean over 
classes averaged over $25$ samples. Whereas Table~\ref{tab:argmax_noisy} shows the results of the distribution 
mode. Our approach outperforms the baseline in both cases, which emphasizes the robustness of our approach 
to noisy input. Nevertheless, the baseline performs slightly better than our approach for the case of 
observing $30\%$ and predicting the next $10\%$ of the videos on the Breakfast dataset. This case corresponds 
to a short-term prediction where in most cases the future activities consist of only one action segment. 
Whereas our approach is better for longer time horizons.

\subsection{Effect of the Number of Samples}

To evaluate the generated samples, we report the mean over classes averaged over $25$ samples. 
Table~\ref{tab:number_of_samples} shows the effect of using different numbers of samples. In each 
case, the average and standard deviation of $5$ runs are reported. As shown in Table~\ref{tab:number_of_samples}, the impact of the number of samples is small. While the average MoC remains in the same range, the 
standard deviation decreases as increasing the number of samples.
\begin{table*}[tb]
\centering
\resizebox{\linewidth}{!}{%
\begin{tabular}{|c|c|c|c|c|c|c|c|c|}
\hline
Observation \% & 
\multicolumn{4}{c|}{\textbf{20\%}} &
\multicolumn{4}{c|}{\textbf{30\%}} \\ \hline
Prediction \%  & 
\textbf{10\%} & \textbf{20\%} & \textbf{30\%} & \textbf{50\%} & 
\textbf{10\%} & \textbf{20\%} & \textbf{30\%} & \textbf{50\%}  \\ \hline

\multicolumn{9}{|l|}{\textbf{\textit{Breakfast}}} \\ \hline
5 samples &
0.1547 $\pm$ 0.0014    &     0.1402 $\pm$ 0.0007     &     0.1349 $\pm$ 0.0020    &     0.1310 $\pm$ 0.0018      &
0.1907 $\pm$ 0.0017    &     0.1703 $\pm$ 0.0016     &     0.1738 $\pm$ 0.0036    &     0.1508 $\pm$ 0.0020     
\\ 

10 samples &
0.1557 $\pm$ 0.0010    &     0.1407 $\pm$ 0.0012     &     0.1347 $\pm$ 0.0017    &     0.1302 $\pm$ 0.0013      &
0.1897 $\pm$ 0.0015    &     0.1708 $\pm$ 0.0012     &     0.1707 $\pm$ 0.0010    &     0.1502 $\pm$ 0.0010    	
\\ 

25 samples &
0.1557 $\pm$ 0.0007    &     0.1406 $\pm$ 0.0008     &     0.1342 $\pm$ 0.0007    &     0.1310 $\pm$ 0.0006      &
0.1904 $\pm$ 0.0006    &     0.1712 $\pm$ 0.0008     &     0.1718 $\pm$ 0.0011    &     0.1509 $\pm$ 0.0006   	
\\ 

50 samples      & 
0.1556 $\pm$ 0.0011    &     0.1400 $\pm$ 0.0006     &     0.1348 $\pm$ 0.0006    &     0.1298 $\pm$ 0.0003      &
0.1906 $\pm$ 0.0011    &     0.1712 $\pm$ 0.0005     &     0.1722 $\pm$ 0.0003    &     0.1507 $\pm$ 0.0006         
\\ \hline

\multicolumn{9}{|l|}{\textbf{\textit{50Salads}}} \\ \hline
5 samples &
0.2404 $\pm$ 0.0200    &     0.2002 $\pm$ 0.0071     &     0.1693 $\pm$ 0.0077    &     0.1247 $\pm$ 0.0078      &
0.2639 $\pm$ 0.0108    &     0.1798 $\pm$ 0.0139     &     0.1453 $\pm$ 0.0118    &     0.1199 $\pm$ 0.0044   	
\\ 

10 samples &
0.2304 $\pm$ 0.0118    &     0.2037 $\pm$ 0.0045     &     0.1739 $\pm$ 0.0054    &     0.1264 $\pm$ 0.0054      &
0.2699 $\pm$ 0.0080    &     0.1807 $\pm$ 0.0075     &     0.1468 $\pm$ 0.0095    &     0.1259 $\pm$ 0.0040  	
\\ 

25 samples &
0.2316 $\pm$ 0.0068    &     0.2003 $\pm$ 0.0040     &     0.1730 $\pm$ 0.0048    &     0.1271 $\pm$ 0.0013      &
0.2658 $\pm$ 0.0042    &     0.1822 $\pm$ 0.0062     &     0.1483 $\pm$ 0.0060    &     0.1220 $\pm$ 0.0029  	
\\ 

50 samples      & 
0.2308 $\pm$ 0.0032    &     0.1998 $\pm$ 0.0010     &     0.1754 $\pm$ 0.0014    &     0.1278 $\pm$ 0.0013      &
0.2664 $\pm$ 0.0030    &     0.1826 $\pm$ 0.0036     &     0.1459 $\pm$ 0.0028    &     0.1225 $\pm$ 0.0037        
\\ \hline

\end{tabular}%
}
\vspace{1mm}
\caption{Effect of the number of samples. Numbers represent mean over classes (MoC) metric averaged 
over the samples. In each case, the average and standard deviation of $5$ runs are reported.}
\label{tab:number_of_samples}

\end{table*}

\subsection{Comparison with the State-of-the-Art}

\begin{table}[tb]
\centering
\resizebox{\columnwidth}{!}{%
\begin{tabular}{|c|c|c|c|c|c|c|c|c|}
\hline
Observation \% & 
\multicolumn{4}{c|}{\textbf{20\%}} &
\multicolumn{4}{c|}{\textbf{30\%}} \\ \hline
Prediction \%  & 
\textbf{10\%} & \textbf{20\%} & \textbf{30\%} & \textbf{50\%} & 
\textbf{10\%} & \textbf{20\%} & \textbf{30\%} & \textbf{50\%}  \\ \hline

\multicolumn{9}{|l|}{\textbf{\textit{Breakfast}}} \\ \hline

RNN model~\cite{abufarha2018when}      & 
0.6035      &       0.5044       &       0.4528      &       0.4042      &
0.6145      &       0.5025       &       0.4490      &       0.4175  
\\ 

CNN model~\cite{abufarha2018when}      & 
0.5797      &       0.4912       &       0.4403      &       0.3926      &
0.6032      &       0.5014       &       0.4518      &       0.4051 	
\\ 

Time-Cond.~\cite{ke2019time}      & 
0.6446      &       0.5627       &       0.5015      &       0.4399      &
0.6595      &       0.5594       &       0.4914      &       0.4423 	
\\ \hline

Ours (Mode)      & 
0.5300      &       0.4410       &       0.3972      &       0.3490      &
0.5399      &       0.4453       &       0.4021      &       0.3558
\\ 

Ours (Top-1)     & 
0.7884      &       0.7284       &     0.6629      &     0.6345      &
0.8200      &       0.7283       &     0.6913      &     0.6239 
\\ \hline

\multicolumn{9}{|l|}{\textbf{\textit{50Salads}}} \\ \hline

RNN model~\cite{abufarha2018when}      & 
0.4230      &       0.3119       &       0.2522      &       0.1682      &
0.4419      &       0.2951       &       0.1996      &       0.1038 			
\\ 

CNN model~\cite{abufarha2018when}      & 
0.3608      &       0.2762       &       0.2143      &       0.1548      &
0.3736      &       0.2478       &       0.2078      &       0.1405 
\\ 

Time-Cond.~\cite{ke2019time}      & 
0.4512      &       0.3323       &       0.2759      &       0.1727      &
0.4640      &       0.3480       &       0.2524      &       0.1384 
\\ \hline

Ours (Mode)     & 
0.3810      &       0.3010       &       0.2633      &       0.1651      &
0.4000      &       0.2927       &       0.2317      &       0.1548
\\ 

Ours (Top-1)     & 
0.7489      &       0.5875       &       0.4607      &       0.3571      &
0.6739      &       0.5237       &       0.4673      &       0.3664  
\\ \hline

\end{tabular}%
}
\vspace{1mm}
\caption{Comparison with the state-of-the-art using ground-truth observations. Numbers represent mean over classes (MoC).}
\label{tab:state_of_the_art_gt}
\end{table}


\begin{table}[tb]
\centering
\resizebox{\columnwidth}{!}{%
\begin{tabular}{|c|c|c|c|c|c|c|c|c|}
\hline
Observation \% & 
\multicolumn{4}{c|}{\textbf{20\%}} &
\multicolumn{4}{c|}{\textbf{30\%}} \\ \hline
Prediction \%  & 
\textbf{10\%} & \textbf{20\%} & \textbf{30\%} & \textbf{50\%} & 
\textbf{10\%} & \textbf{20\%} & \textbf{30\%} & \textbf{50\%}  \\ \hline

\multicolumn{9}{|l|}{\textbf{\textit{Breakfast}}} \\ \hline

RNN model~\cite{abufarha2018when}      & 
0.1811       &      0.1720  &   0.1594  &  0.1581  &
0.2164       &      0.2002   &  0.1973  &  0.1921  
\\ 

CNN model~\cite{abufarha2018when}      & 
0.1790	   &	0.1635		&	0.1537			&	0.1454			&
0.2244	   &	0.2012  	&	0.1969			&	0.1876		
\\ 

Time-Cond.~\cite{ke2019time}      & 
0.1841      &       0.1721       &       0.1642      &       0.1584      &
0.2275      &       0.2044       &       0.1964      &       0.1975 
\\ \hline

Ours (Mode)     & 
0.1671       &       0.1540       &       0.1447      &       0.1420      &
0.2073       &       0.1827       &       0.1842      &       0.1686  
\\ 

Ours (Top-1)     & 
0.2889      &       0.2843       &       0.2761      &       0.2804      &
0.3238      &       0.3160       &       0.3283      &       0.3079  
\\ \hline

\multicolumn{9}{|l|}{\textbf{\textit{50Salads}}} \\ \hline

RNN model~\cite{abufarha2018when}      & 
0.3006	&	0.2543      	&	0.1874       	&	0.1349	&
0.3077	&	0.1719			&	0.1479			&	0.0977			
\\ 

CNN model~\cite{abufarha2018when}      & 
0.2124			&	0.1903		  &	    0.1598		  & 	0.0987			&
0.2914			&	0.2014	      &	    0.1746	      &	    0.1086
\\ 

Time-Cond.~\cite{ke2019time}      & 
0.3251      &       0.2761       &       0.2126      &       0.1599      &
0.3512      &       0.2705       &       0.2205      &       0.1559 
\\ \hline

Ours (Mode)     & 
0.2486       &       0.2237       &       0.1988      &       0.1282      &
0.2910       &       0.2050       &       0.1528      &       0.1231   
\\ 

Ours (Top-1)     & 
0.5353      &     0.4299       &     0.4050      &       0.3370      &
0.5643      &     0.4282       &     0.3580      &       0.3022     
\\ \hline

\end{tabular}%
}
\vspace{1mm}
\caption{Comparison with the state-of-the-art without ground-truth observations. Numbers represent mean over classes (MoC).}
\label{tab:state_of_the_art_noisy}
\end{table}

\begin{table}[tb]
\centering
\resizebox{.5\columnwidth}{!}{%
\begin{tabular}{c|c}
\hline
Model & Accuracy \\
\hline

APP-VAE~\cite{mehrasa2019variational}      &  \textbf{62.2}
\\ 

Ours      &  57.8
\\ \hline

\end{tabular}%
}
\vspace{1mm}
\caption{Comparison with \cite{mehrasa2019variational}: Accuracy of predicting the label 
of the next action segment.}
\label{tab:state_of_the_art_accuracy}
\end{table}

In this section, we compare our approach with the state-of-the-art methods for anticipating activities. 
Since the state-of-the-art methods do not model uncertainty and predict just a single sequence of future 
activities, we report the accuracy of the mode of the predicted distribution of our approach. 
Table~\ref{tab:state_of_the_art_gt} shows 
the results with the ground-truth observations, and Table~\ref{tab:state_of_the_art_noisy} shows the 
results without the ground-truth observations. While the mode of the distribution from our approach only 
outperforms the CNN model~\cite{abufarha2018when} on the 50Salads dataset, it shows a lower accuracy compared 
to the RNN model and~\cite{ke2019time}. This is expected since these approaches were trained to predict only 
a single sequence of future activities while our approach is trained to predict multiple sequences. We therefore 
also report the top-1 MoC. The results show that our model captures the future activities better.

We also compare our approach with~\cite{mehrasa2019variational} in predicting the next action segment given 
all the previous segments. The accuracy of predicting the label of the future action segment is reported in 
Table~\ref{tab:state_of_the_art_accuracy}. Note that in this comparison we use the ground-truth annotations 
of the videos as in~\cite{mehrasa2019variational}. Our approach achieves a lower accuracy. However, our approach 
is designed for long-term prediction as we have already observed in Section~\ref{sec:anticipation_without_gt_obs} 
and this comparison considers short-term prediction only. Moreover, the approach of~\cite{mehrasa2019variational} 
is very expensive, making it infeasible for long-term predictions.


\section{Conclusion}
\label{sec:conclusion}
We presented a framework for modelling the uncertainty of future activities. Both an action 
model and a length model are trained to predict a probability distribution over the future action 
segments. At test time, we used the predicted distribution to generate many samples. Our framework 
is able to generate a diverse set of samples that correspond to the different plausible future 
activities. While our approach achieves comparable results for short-term prediction, our approach 
is in particular useful for long-term prediction since for such scenarios the uncertainty in the 
future activities increases.

\paragraph{Acknowledgements:} 
The work has been funded by the Deutsche Forschungsgemeinschaft (DFG, German Research Foundation) – GA 1927/4-1 (FOR 2535 Anticipating Human Behavior) and the ERC Starting Grant ARCA (677650).

{\small
\bibliographystyle{ieee_fullname}
\bibliography{egbib}
}

\end{document}